# Deriving Contextualised Semantic Features from BERT (and Other Transformer Model) Embeddings

Jacob Turton[1], David Vinson[2],
Robert Elliott Smith[1]
[1]Department of Computer Science, UCL
[2]Department of Psychology and Language Sciences, UCL

## Abstract

Models based on the transformer architecture, such as BERT, have marked a crucial step forward in the field of Natural Language Processing. Importantly, they allow the creation of word embeddings that capture important semantic information about words in context. However, as single entities, these embeddings are difficult to interpret and the models used to create them have been described as opaque. Binder and colleagues proposed an intuitive embedding space where each dimension is based on one of 65 core semantic features. Unfortunately, the space only exists for a small dataset of 535 words, limiting its uses. Previous work (Utsumi, 2018, 2020, Turton, Vinson & Smith, 2020) has shown that Binder features can be derived from static embeddings and successfully extrapolated to a large new vocabulary. Taking the next step, this paper demonstrates that Binder features can be derived from the BERT embedding space. This provides contextualised Binder embeddings, which can aid in understanding semantic differences between words in context. It additionally provides insights into how semantic features are represented across the different layers of the BERT model.

## 1   Introduction

The last decade or so has seen a rapid progress in the field of Natural Language Processing (NLP) with a combination of new models and increasingly powerful hardware resulting in state of the art performances across a number of common tasks (Wang et al, 2020). One important area of improvement has been in the vector-space representation of words, known as word embeddings. Embedding models create word vectors within a vector space that captures important semantic and grammatical information (Boleda, 2020). Distributional models based on neural architectures, such as Word2Vec (Mikolov et al, 2013) and GloVe (Pennington, Socher & Manning, 2014) became popular in the 2010s, but only produce a single static embedding for each word in a vocabulary. In reality, words can have multiple meanings, whether completely different (homonyms) or different but still somewhat related (polysemy). Static models can only "learn" one embedding per word form, which is limiting, as roughly 7% of *common* English word forms have homonyms and over 80% are polysemous (Rodd et al, 2002).

ELMO: Embeddings from Language Models (Peters et al, 2018), was the first major model to tackle this problem by creating contextualised embeddings using the hidden states from a bidirectional Long Short-Term Memory (LSTM) recurrent network. Following this, the introduction of the transformer architecture and in particular its implementation in the Bidirectional Encoder Representations from Transformers (BERT) (Devlin et al, 2018) model, resulted in even better contextual embeddings. BERT was soon followed by a flurry of models which used its general architecture, but made important improvements to its pre-training methods (e.g. Liu et al, 2019, Yang et al, 2019) further advancing the state of the art.

Regardless of whether the embeddings mentioned are static or contextual, they all have the issue that, as individual objects, they are hard to interpret (Şenel et al, 2018). Whilst efforts have been made to produce more interpretable embeddings (e.g. Panigrahi et al, 2019), the general approach has been to interpret them in relation to each-other. For example, the relative distance between word embeddings can indicate their semantic similarity (Schnabel et al, 2015). Alternatively, using dimensionality reduction methods such as Principal Components Analysis (PCA) or T-distributed Stochastic Neighbor Embedding (t-SNE) to project the embeddings into a two or three dimensional space allows their relative positions to be visualised (Liu et al, 2017). However, these methods may just show *how* the embeddings are related, rather than *why*. This issue further feeds into the general criticism levelled at deep learning architectures, such as BERT, that they are opaque and difficult to interpret (Belinkov and Glass, 2019).

Binder and colleagues (2016) presented an alternative embedding space for words, based on 65 core semantic features, where each dimension relates to a feature. Unfortunately, the Binder dataset only contains 535 words as they were scored by human participants, severely limiting its use for large scale text analysis. However, previous research (Utsumi, 2018, 2020, Turton, Vinson & Smith, 2020) has shown that the Binder feature values can be derived from static embeddings, such as Word2Vec, and successfully extrapolated to a large new vocabulary of words. The purpose of this research is to demonstrate that Binder features can be successfully derived from BERT embeddings, allowing contextualised semantic features to be predicted for any word in context.

## 2 Related Work

### 2.1 Bidirectional Encoder Representations from Transformers (BERT)

The Bidirectional Encoder Representations from Transformers (BERT) language model (Devlin et al, 2018) uses a bidirectional transformer architecture. The model is trained using two tasks. The first, masked language modeling, involves predicting randomly masked words (15% of total) in a sequence. The second, next sentence prediction, detects whether two sentences are consecutive or not. Together with its transformer architecture, the pre-training allows BERT to learn the complex interactions between words in a sequence. Importantly, contextualised word embeddings can be retrieved from the BERT model by extracting its internal hidden states. However, since the BERT model has a number of layers (12 in $BERT_{BASE}$ and 24 in $BERT_{LARGE}$) any of these can be extracted as a contextual embedding and therefore it is important to find the best representation for the task at hand. Probing tasks, discussed later, offer some insight into this.

### 2.2 Other Transformer Models

Following the release of BERT, a number of models using the same general architecture, but slightly altered pre-training methods, were released. XLNet (Yang et al, 2019) used permutation instead of masked language modeling. RoBERTA (Liu et al, 2019) altered pre-training by removing next sentence prediction, randomising the masked words between epochs and using a much larger training dataset. These changes lead to both models outperforming BERT on a number of important benchmark tasks. OpenAI's GPT (Radford et al, 2019) was actually released before BERT and uses a unidirectional transformer architecture. The follow-up GPT-2 model relied on size and large amounts of training data to drive performance whilst retaining the unidirectional architecture.

### 2.3 Probing Language Models

Whilst the models discussed so far have led to impressive improvements in NLP tasks, they have also been criticised as opaque and difficult to interpret (Castelvecchi, 2016). Therefore, researchers have made efforts to better understand how they work. For example, Clark et al (2019) were able to show that patterns of attention in BERT respond to certain syntactic relations between words. Other work has looked at how semantic information in represented in BERT. Researchers have shown that BERT can learn to represent semantic roles (Ettinger, 2019), entity types and semantic relations (Tenney et al, 2019). Reif et al (2019) demonstrated clear 'clusters' for different senses of the same word, when visualising the spatial location of their BERT embeddings. Other researchers have shown that the embeddings from different layers of BERT perform better at different tasks. Jawahar et al (2019) found that certain surface features, such as sentence length were better encoded in earlier layers, whereas syntac-tic information tended to be better represented by middle layers and semantic information by the later layers.

### 2.4 Binder Semantic Features

Through a meta-analysis, Binder et al (2016) identified 65 semantic features all believed to, and some demonstrated to, have neural correlates within the brain. They then asked participants to rate a total of 535 words on a scale of 0-6 for each of the features. The features ranged from concrete object properties such as visual and auditory, to more abstract properties such as emotional aspects. This resulted in a 65-dimensional embedding for each word, where each dimension relates to a specific semantic feature.

This embedding space is useful as each dimension is easily interpretable and theoretically connected to a specific aspect of how people understand the meaning of words and concepts. Furthermore, representing words in this way makes it easy to understand how they are similar or different in terms of their semantic features. Figure 1 below demonstrates this by comparing the feature scores of the words *raspberry* and *mosquito*. It shows how the concepts differ across a range of dimensions. Also, since these features are derived from the psychological and neuroscience literature, it may mirror how people differentiate these concepts.

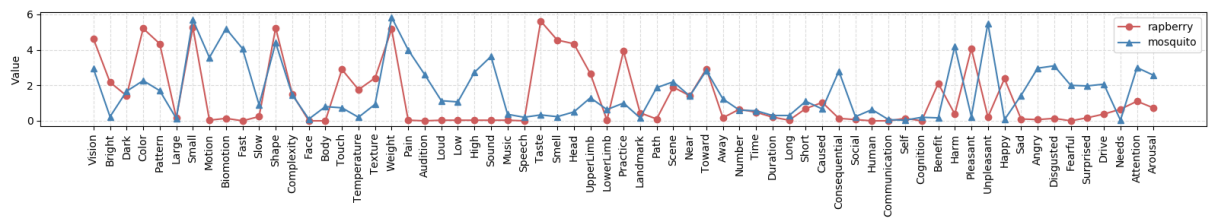

Figure 1. Feature values for the words *raspberry* and *mosquito* in comparison

Unfortunately, the Binder dataset only exists for 535 words, which severely limits its uses. Since human raters are used to create the embeddings, it would be prohibitively expensive to expand it to a decently sized vocabulary useful for any large scale text analytics. Previous work (Utsumi, 2018) has shown that Binder feature vectors can be derived from dense word embeddings such as Word2Vec and that this can be used to extrapolate the feature space to a large number of new words (Turton, Vinson & Smith, 2020). Being able to do this using BERT embeddings would allow the features to be derived for words in context. Not only would this tackle the issue of polysemy, but hopefully also mirror more subtle differences between words when used in context. Beyond this, the dataset also offers a powerful way to probe the semantic representation of words in models like BERT by looking at how well the different semantic features can be predicted overall, how the semantic representations build over the layers of the models and whether there are distinct patterns in how different types of semantic feature are represented across the layers.

## 3 Experiment 1a: Deriving Binder Embeddings from BERT and other Transformer Model Embeddings

### 3.1 Introduction

The purpose of this first experiment was to demonstrate that it is possible to derive Binder style word embeddings primarily from the BERT embedding space (but also a select few additional models for comparison). Since the Binder dataset words are provided out of any context, the BERT embeddings were treated as static by taking 250 random sentences containing each target word and averaging the embedding across all 250. The Binder features for the words could then be predicted from these 'static' embeddings. For comparison, embeddings from RoBERTa, XLNet and GPT-2 were also included. Numberbatch static embeddings (Speer, Chin & Havasi, 2017), the best performing static embeddings seen previously (Turton, Vinson & Smith, 2020), were used as a baseline.

Since it was not known which layer of the BERT model would provide the best embeddings for this task, all were evaluated separately. This also offered the opportunity to investigate how the different semantic features are represented across the different layers in the BERT model. Binder et al

(2016) categorised the different semantic features into broad domains, such as Vision and Audition. It is possible that these categories of features may be mirrored in how the features are represented across the layers, which would suggest BERT captures higher level human semantic understanding.

### 3.2 Materials

The Binder et al (2016) dataset was used providing scores across the 65 features for 535 words. For random sentences containing the Binder words, the One Billion Word Benchmark (BWB) (Chelba et al, 2013) was used. Pre-trained BERT, XLNET, RoBERTa & GPT-2 models were used, with both the base and large versions compared. As far as possible, models of the same size were selected (see Appendix Table i for further details). Pre-trained Numberbatch embeddings were also used (Speer, Chin & Havasi, 2017). A simple 3 hidden-layer neural network was used to predict the semantic feature values from the embeddings.

### 3.3 Method

**The method here describes the process for the BERT$_{BASE}$ model, but was the same for all other models as well**.

To get "static" embeddings for each of the 535 Binder target words, 250 random sentences containing each of the target words were selected from the BWB dataset. Using the pre-trained BERT$_{BASE}$ model the embeddings from all 12/24 layers and the embedding layer were extracted for the target word for each of the sentences. A mean of the target word embedding across the sentences was taken. Additionally, for each model the best performing subword embedding approach was used (see Table ii and Figure i in Appendix for comparisons).

A 3-hidden-layer neural network model was used to to predict the semantic feature values. The extracted embeddings were used as the input and each of the 65 semantic features were predicted separately. Since the Binder dataset is small (535 words) and the dimensionality of the inputs large (768-1024), it was important to maximise the training sample size to avoid overfitting. However, it was also important to ensure that the model was evaluated on a validation set that was sufficiently diverse. Therefore, 20-fold cross validation was used. This ensured a large training sample each time (95%) and the models were ultimately validated across the entire set of Binder words. For each semantic feature, the average R-squared score from across all of the folds was calculated as the final score.

To investigate how the different semantic features are represented across the layers, each feature's r-squared score was rescaled between 0-1 across the layers. A k-means clustering algorithm was then used to group the features according to similar patterns across the layers. The rescaling ensured it was the pattern of behaviour across the layers rather than the absolute performance of each feature that was captured in the clustering. The membership of the clusters was compared to the categories of the features given in Binder et al (2016) using the Adjusted Rand Index (Yeung & Ruzzo, 2001). Wilcoxon Ranks-sums test (Demsar, 2006) was used to compare performance of the different embedding models.

### 3.4 Results

Figure 2 below shows the mean R-squared scores across all semantic features for the different layers for the large and small models. The models generally show a rise in performance as the layer rises followed by a drop off, although with slightly different patterns. For example XLNet peaks much earlier than BERT. GPT-2 clearly has the worst performance overall in terms of both the small and large models. Table 1 reiterates these findings in its second row. The first row of Table 1 shows the overall mean R-squared scores for each of the models, but this time taking the score for each feature from its best performing layer. All models perform better using this selective process. All models except GPT-2$_{SMALL}$ significantly outperformed Numberbatch from a Wilcoxon Signed Ranks test (p < 0.05 for all). BERT$_{BASE}$ also outperformed XLNet$_{BASE}$ (p<0.05) but not RoBERTa$_{BASE}$ (p=0.17).

There was substantial variation between the features in how well they were predicted, with some relatively low (e.g. Slow ~0.3) and others higher (e.g. Shape ~0.8) (See Figure ii in the Appendix for full results). There was also general consistency between the models as to which features were well and poorly predicted with inter-feature variance (mean=0.011) larger than inter-model variance (mean=0.001). This indicates certain semantic features are difficult to predict regardless of the model.

For all models the larger version performed significantly better than the base version (p<0.05 for all). For the larger models there was no longer any significant difference between the BERT$_{LARGE}$, RoBERTa$_{LARGE}$ and XLNet$_{LARGE}$ models (p > 0.05 for all), but all three did outperform GPT-2$_{MEDIUM}$ (p>0.05 for all).

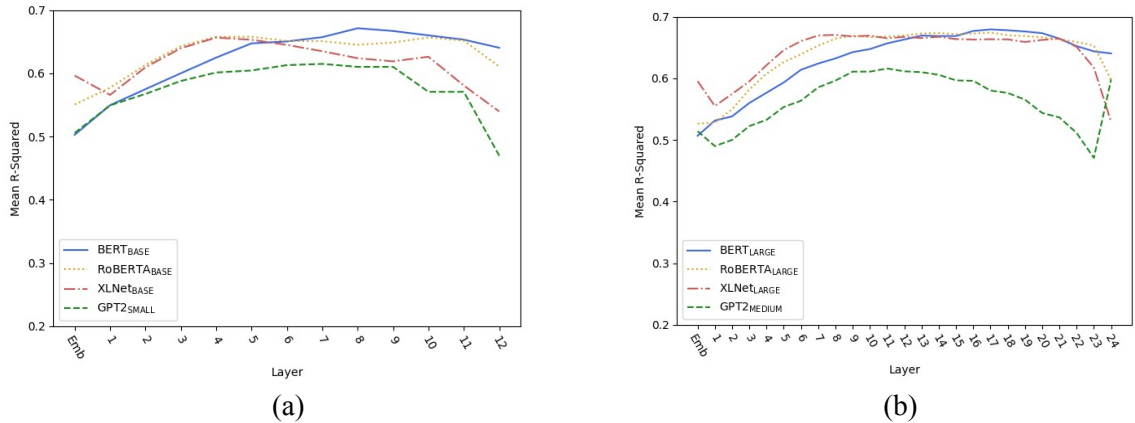

Figure 2. Mean R-squared scores across all semantic features for layers of (a) small and (b) large models.

| MEAN R-SQUARED | NmbrBatch | MODEL | | | | | | | |
|---|---|---|---|---|---|---|---|---|---|
| | | GPT-2 | | Roberta | | XL-Net | | BERT | |
| | | *Small* | *Med.* | *Base* | *Large* | *Base* | *Large* | *Base* | *Large* |
| Any Combined | - | .631 | .638 | .673 | **.692** | .665 | .688 | **.678** | .692 |
| Best Layer Only | .646 | .615 | .616 | .658 | .674 | .656 | .670 | **.667** | **.679** |

Table 1. Best overall mean R-squared scores for the models across all 65 semantic features (best layer in brackets).

The k-means clustering on the re-scaled BERT$_{BASE}$ R-squared scores indicated an optimal 3 clusters identified using an elbow plot. Figure 3 (a) below shows the memberships of the k-means clusters, along with their respective mean scores across each layer (Appendix figure iii for full results). Cluster 0 and 1 show a similar pattern showing a peak in the later layers. Cluster 2 shows a very different pattern with the peak much earlier in the mid-layers. Figure 3 (b) shows the mean raw R-squared layer scores for the different clusters. Clusters 0 & 2 have similarly high peaks whereas cluster 1's peak is much lower. Whilst this does suggest different patterns of representation for the different features in the model, the clusters do not appear to match the categorisation of features given by Binder et al (2016) as the adjusted rand index was 0.02, very close to 0 indicating similar performance to random.

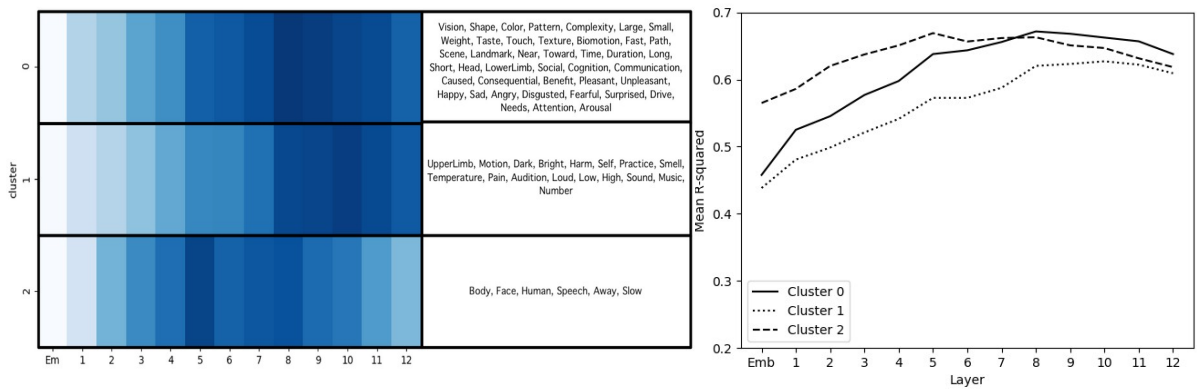

Figure 3. (a) mean re-scaled R-squared scores for the three clusters with member features and (b) mean layer raw R-squared scores for the three clusters.

## 3.5 Discussion

The main purpose of this first experiment was to demonstrate that Binder style embeddings can successfully be derived from the BERT (and other similar model) embedding space. The secondary purpose was to explore how the representation of the semantic features varies across the different layers of a $BERT_{BASE}$ model. The results demonstrated that Binder features could be derived from BERT embeddings, outperforming static Numberbatch embeddings. This is interesting as Numberbatch embeddings make use of additional human provided information from a concept network, whereas BERT and the other models are purely trained on raw text. This hints towards the power of these bidirectional transformer models in capturing semantic information from word usage.

Whilst RoBERTa and XLNet have both been shown to outperform BERT on a number of NLP tasks, they did not outperform BERT here, with BERT actually performing slightly better than both in the base models and roughly equal to RoBERTa in the large models. The poor performance of GPT-2 in this case is not surprising as it was limited to a relatively small size (for it) to match with the other models in terms of total parameters. GPT-2 has shown success when using very large models (up to 1.5B parameters, compared to $BERT_{LARGE}$'s 340M). These results highlight the power of the bidirectional architecture used by BERT, XLNet and RoBERTa.

Perhaps most interesting results from this experiment are in relation to how the different semantic features are represented across the layers of BERT. In line with the findings of Jawahar et al (2019), semantic features tended to be better represented by the later layers. However, a small subset of features were better represented by the middle layers. Clustering the features according to these behaviours did not match the Binder categories. However, Binder et al's categorisation of the features is not the only way to group them and there are some similarities between the features in the different clusters. For example, Cluster 3 appears to capture a number of features (Human, Face, Speech, Body) relating to people and Cluster 2 captures 6 of the 7 features relating to audition.

## 4 Experiment 1b: Towards Contextualised Binder Features

### 4.1 Introduction

Experiment 1a demonstrated that Binder semantic features can be predicted from the BERT (and other model) embedding space, outperforming the best performing static embeddings (Numberbatch). However, the real power of the transformer architecture and its self-attention mechanism, is being able to represent a contextualised form of a word token (Reif et al, 2019). By treating the embeddings as "static" and using randomly selected sentences for Experiment 1a, it was limited to an average embedding of the word over many contexts. This may have reduced performance by including uses of the word that did not match the sense of the word suggested by the Binder feature scores. Instead, it is possible to infer the sense for each word in the Binder dataset from its feature scores. For example the scores for *arm* suggest the human limb sense of the word, rather than a weapon. Selecting only sentences that use the words in the sense inferred from their feature scores should improve performance as it removes unrelated word uses form the averaged embedding. Additionally, it begins to demonstrate how the transformer architecture can capture the different features of a word depending on context.

### 4.2 Material

The same materials as were used as Experiment 1a with ten "hand-picked" sentences from the BWB per word rather than 200 randomly selected ones.

### 4.3 Method

For each word in the Binder dataset, ten sentences were hand-picked from the 200 previously randomly selected BWB sentences. Sentences were picked by matching them to the word sense inferred from the Binder feature scores. Following this, the exact same method as Experiment 1a was used, using the average embedding across the ten hand-selected sentences rather than the 200 randomly selected ones. Only the larger versions of the models were used as they performed best in Experiment 1a.

### 4.4 Results

Table 2 below gives the mean results for the large version of BERT, RoBERTa and XL-Net and the medium version of GTP-2 using the focused sentences with BERT from Experiment 1a as a baseline. (Per feature results can be found in Figure iv of the Appendix)

| MEAN R-SQUARED | BASELINE | | MODEL GPT-2 | | RoBERTa | | XL-Net | | BERT | |
|---|---|---|---|---|---|---|---|---|---|---|
| | *Base* | *Large* | *Small* | *Med.* | *Base* | *Large* | *Base* | *Large* | *Base* | *Large* |
| Combined | .678 | .692 | .656 | .670 | **.736** | **.755** | .707 | .730 | .725 | .741 |
| Best Layer | .667 | .679 | .638 | .643 | **.723** | **.741** | .697 | .714 | .718 | .729 |

Table 2. Mean R-squared scores for the models using selected sentences vs BERT baseline from Experiment 1a (randomly selected sentences)

As Table 2 shows, except GPT-2, all embeddings from Experiment 1b outperform the BERT baseline from Experiment 1a, indicating that using the selected sentences rather than randomly sampled sentences gives better performance.

### 4.5 Discussion

Using hand selected rather than purely randomly selected sentences improved the performance as expected. This was likely due to removing noise from unrelated uses of the word in the averaged embedding. Whilst this demonstrates that context plays a role in predicting Binder feature scores for words, it still falls short of a ground-truth test for the ability to derive Binder feature scores for words in specific contexts. The Binder dataset does not provide contexts for its words and the participants in their experiment were not asked rate words in a specific context. Therefore, to investigate how well semantic features can be predicted for words in specific contexts, it is necessary to look at other datasets.

## 5 Experiment 2: Predicting Explicit Contextualised Features

### 5.1 Introduction

Together Experiments 1a & b demonstrate that semantic features ratings can be derived from transformer embeddings and that introducing some degree of context improves the performance. But the Binder dataset unfortunately lacks explicit context for the rated words.

An alternative dataset (van Dantzig et al, 2011) of contextualised semantic features for words exists with words in context pairs. In each context pair a property word e.g. *abrasive* is paired an object e.g. *lava* and participants scored the property word across five semantic features. Each property is paired with two different objects giving two word-pairs for each property. By feeding the property-object pairs into the transformer models, the extracted embedding for the property word should capture the its specific feature values influenced by its context object word.

As a comparison, Numberbatch static embeddings were also used to predict the feature values. The transformer embeddings should outperform the static Numberbatch embeddings which are unable to account for the context of the object paired with the property. In case it is simply the transformer embeddings outperforming the Numberbatch embedding without necessarily capturing any context, the mean property word embeddings between both of the property-object pairs was also evaluated. The contextual transformer embeddings should outperform these mean embeddings if the models are able to capture the specific property-object contextual semantic features of the property words.

Due to its poor performance in Experiments 1a & b, the GPT-2 model was no longer considered and only the better performing $_{LARGE}$ versions of BERT, Xl-Net and RoBERTa were used.

### 5.2 Materials

The word-pair dataset (van Dantzig et al, 2011) consists of 774 property-object pairs. Each word pair consists of a property and object word, and has a rating across the semantic features: *Visual*, *Auditory*, *Haptic*, *Gustatory* and *Olfactory*. The ratings are between 0-5 for each. Table 3 below gives an example of the first two property words in their word-pairs with their ratings. The same pretrained BERT$_{LARGE}$, XL-Net$_{LARGE}$ and ROBERTA$_{LARGE}$ models from Experiment 1a/b were used and the pre-trained embeddings for Numberbatch.

| | | FEATURE | | | | |
|---|---|---|---|---|---|---|
| **PROPERTY** | **OBJECT** | Visual | Auditory | Haptic | Gustatory | Olfactory |
| Abrasive | Lava | 3.83 | 1.27 | 2.37 | 0.07 | 0.46 |
| Abrasive | Sandpaper | 3.37 | 2.35 | 4.81 | 0.26 | 0.09 |
| Babbling | Baby | 3.71 | 4.83 | 0.78 | 0.00 | 0.24 |
| Babbling | Brook | 3.08 | 4.00 | 1.14 | 0.19 | 0.22 |

Table 3. Semantic feature values for the property words *abrasive* and *babbling* with different objects

## 5.3 Method

The property-object word pairs were fed into the transformer models as the input sequences and the embedding for the property word was extracted. Embeddings from all 24 layers (layers 1-24) and the embedding layer (layer 0) were extracted. The embeddings were then fed into a simple 3-layer neural network for training & prediction with each of the five semantic features used separately as the target variable.

Like Experiment 1, the dataset was split into ten-folds with 90% of the data for training and the reaming 10% for evaluation. The mean r-squared scores across the ten-folds was calculated for each of the five semantic features.

## 5.4 Results

Table 4 below shows the performance of embeddings from the different models for predicting the five features. A Numberbatch baseline is given and for each of the models. The best performing layer for each of the features from each of the models is selected. (See Appendix Figure v for per layer results)

| FEATURE | BASELINE | MEAN | | | CONTEXTUALISED | | |
|---|---|---|---|---|---|---|---|
| | Numberbatch | BERT | XL-Net | RoBERTa | BERT | XL-Net | RoBERTa |
| Visual | 0.377 | 0.532 | 0.448 | 0.456 | **0.652** | 0.583 | 0.633 |
| Auditory | 0.651 | 0.722 | 0.668 | 0.680 | **0.793** | 0.733 | 0.772 |
| Haptic | 0.516 | 0.556 | 0.512 | 0.505 | **0.660** | 0.616 | 0.634 |
| Gustatory | 0.428 | 0.611 | 0.531 | 0.591 | 0.800 | 0.704 | **0.813** |
| Olfactory | 0.377 | 0.610 | 0.587 | 0.597 | **0.740** | 0.736 | 0.731 |
| **MEAN** | 0.523 | 0.607 | 0.549 | 0.556 | **0.729** | 0.674 | 0.717 |

Table 4. Mean R-squared scores for the five features for mean and contextualised embeddings from the three different models, compared to a Numberbatch baseline.

As Table 4 shows, the contextualised transformer embeddings outperform both the mean transformer embeddings and baseline Numberbatch embeddings. Overall, the BERT model performed best.

## 5.5 Discussion

The purpose of experiment 2 was demonstrate the ability to derive contextual semantic features from transformer embeddings. As predicted, the results show that the contextual transformer embeddings performed better than static Numberbatch embeddings at predicting semantic feature values. Furthermore, the contextual embeddings performed better than the mean transformer embeddings, suggesting it was not just a case of the transformer embedding spaces performing better without picking up the specific contextual information. The mean embeddings outperforming Numberbatch could be due to two factors. First, as experiment 1 demonstrated, transformer embeddings perform slightly better at predicting semantic features than Numberbatch. Second, the mean transformer embeddings are only 'static' across the two property-object pairs, whereas the Numberbatch embeddings are static across the entire text dataset they were trained on. Therefore even though the mean embeddings do not capture the exact context of each property word, it has a higher weighting than in Numberbatch embeddings.

Whilst this experiment begins to show the ability to derive contextualised semantic features from transformer embeddings, it was only done for a very small number of features and trained on word-pairs rather than longer sequences. Ideally, we would be able to predict the full 65 semantic features in the Binder embedding space for words contextualised in longer sequences.

## 6 Experiment 3: Evaluation of Contextualised Binder Embeddings

### 6.1 Introduction

Experiment 1a & b demonstrated that Binder features can be derived from various transformer embedding spaces and that some effects of context can be picked up, whilst Experiment 2 demonstrated that they are effective at predicting contextualised semantic features, but for a very limited number of features and only for word-pair contexts. The lingering issue from these experiments is that a ground-truth dataset for Binder semantic features of words in context is missing, which makes it difficult to evaluate how well fully-contextualised Binder embeddings can be derived from transformer model embeddings.

To address this, this experiment uses the word-sense disambiguation (WSD) task to evaluate derived contextualised Binder embeddings. Whilst it is not a direct measure of model performance in predicting Binder features for contextualised words, it is an indirect measure through the "quality" of the embeddings. WSD is an open problem in NLP where the task is to determine which sense of word is being used in a sequence (Navigli, 2009). Models that perform well on this task are able to separate the different semantic meaning of a word, depending on he context it is used in.

By evaluating how well derived Binder embeddings perform at this task, it should indicate how good the embeddings are at representing the contextualised semantic features of the words. In this experiment the binder embeddings are compared to the raw BERT embeddings from which they are derived.

### 6.2 Materials

The Word in Context (WiC) dataset (Pilehvar & Camacho-Collados, 2019) was used to test derived Binder embeddings on WSD. The dataset consists of a sentence pair each containing the same target word and a binary classification (True/False) of whether the target word has the same word-sense or not between the sentences. The dataset is already divided into a training and separate development set which was used for the final evaluation.

Pretrained $BERT_{LARGE}$ was used to produce word embeddings and the trained neural networks from Experiment 1b were used to predict Binder feature values from them.

### 6.3 Method

Using the pre-trained $BERT_{LARGE}$ model, word embeddings from all 24 layers + the embedding layer were extracted for the target word in each of the sentences of the WiC dataset. Using the neural networks trained in Experiment 1b, the Binder features were predicted using the optimal $BERT_{LARGE}$ layer for each of the 65 features. For each sentence pair, the cosine similarity was calculated between the embeddings for the target words, either using the raw $BERT_{LARGE}$ embeddings or the derived Binder embeddings. For the raw embeddings, the cosine similarity was calculated for the embeddings from all 24 layers separately.

To evaluate the embeddings, a simple logistic regression model was used with the cosine similarity between the embeddings as the input variable. The logistic regression model was trained on the training set and Accuracy and F1 scores were calculated on the dev set for all the embeddings.

### 6.4 Results

Table 5 below shows the performance of the best performing layer (21) raw $BERT_{LARGE}$ embeddings and Binder embeddings on the WiC dev set (see Appendix Figure vi for all layer performances).

| METRIC | Raw $BERT_{LARGE}$ | BERT Binder |
|---|---|---|
| Accuracy | 0.68 | 0.67 |
| F1 Score | 0.71 | 0.71 |

Table 5. Accuracy and F1 score of raw BERT and BERT-derived Binder embeddings on the dev set of WiC.

As can be seen in Table 5, Binder embeddings perform comparatively to the raw BERT$_{LARGE}$ embeddings.

### 6.5 Discussion

The purpose of this final experiment was to evaluate contextualised Binder embeddings. In the absence of a ground-truth dataset for contextualised Binder features, the WSD task was used as an indirect measure. The contextualised Binder embeddings performed comparatively to raw BERT embeddings which are known to capture contextualised semantic meaning of words (Reif et al, 2019, Pilehvar & Camacho-Collados, 2019). This suggests that the contextualised Binder embeddings derived from the BERT$_{LARGE}$ model are able to capture the semantic features of words in specific contexts, within the bounds of the WSD task.

Importantly, the nature of the Binder feature space makes interpreting the embeddings fairly easy helping identify where the embeddings do a good or poor job at representing words in context. For example, Figure 5 below illustrates an example where the derived Binder embeddings do a good job of capturing the different senses of a word in two different contexts for the WiC dataset.

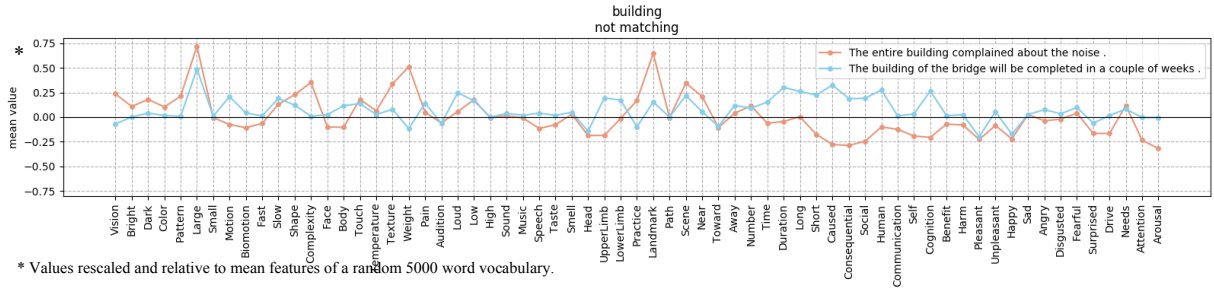

Figure 5. Predicted Binder feature values for the word *building* in different contexts.

Alternatively, Figure 6 gives an example of poorer performance by derived Binder embeddings. It appears the second target word *caught* is more similar to the physical act of catching, as seen by the higher scores for Bio-motion and Body, rather than catching fire.

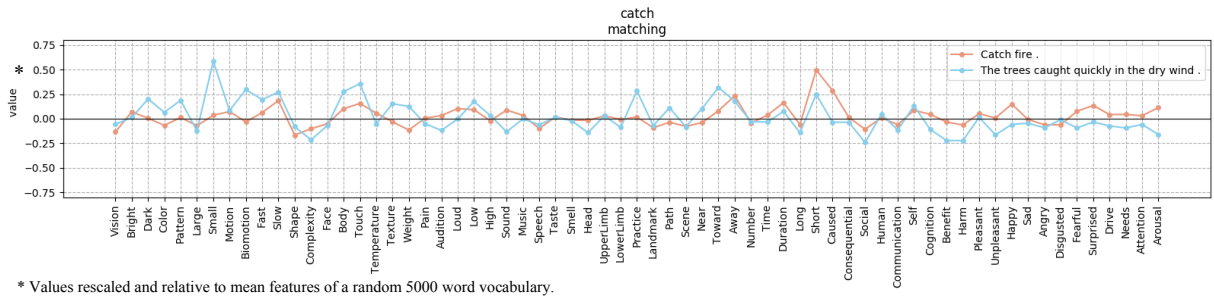

Figure 6. An example of the predicted Binder features indicating different senses of a word, where human raters scored them the same.

## 7 Conclusion

The overarching aim of this work was to demonstrate that Binder style semantic feature embeddings can be derived from the BERT embedding space in the same way that previous research (Turton, Vinson & Smith, 2020, Utsumi, 2018, 2020) has shown they can be from static embeddings. This would allow contextualised Binder embeddings to be derived for new words. It also offered to opportunity to probe how semantic information is represented across the different layers of the BERT model. The first experiment (1a) demonstrated that Binder features can be derived from the BERT embedding space, significantly outperforming the previously best performing static embeddings (Numberbatch). Whilst semantic information was generally better represented by the later layers, there was some variation and certain features were actually better represented by the middle layers. Experiment 1b demonstrated that by selecting sentences to match the suspected word-sense in the Binder dataset, the performance could be improved, providing some evidence that embeddings extracted from transformer architectures capture the contextual semantics of words. Experiment 2 further supported the ability of the transformer architecture to capture contextual semantics, but for a very small set of semantic features and only for word-pair contexts. In the absence of a ground-truth dataset with the full 65 Binder features for words in specific contexts, the WiC WSD dataset was used to test the quality of contextualised Binder embeddings. The Binder embeddings performed comparatively to raw BERT embeddings suggesting they represent the contextualised state of words, at least to the same extent as raw BERT embeddings. Importantly, as demonstrated in Experiment 3, deriving Binder feature scores for words in context, allows them to presented in an embedding space that is easy to interpret and allows the quick and easy comparison of words (as seen in Figures 5 & 6).

# Appendix

|  | MODEL | | | | | | | |
|---|---|---|---|---|---|---|---|---|
|  | BERT | | GPT-2 | | XLNet | | RoBERTA | |
|  | *Base* | *Large* | *Small* | *Medium* | *Base* | *Large* | *Base* | *Large* |
| Parameters | 110M | 340M | 117M | 345M | 110M | 340M | 125M | 355M |
| Layers | 12 | 24 | 12 | 24 | 12 | 24 | 12 | 24 |
| Attention Heads | 12 | 16 | 12 | 16 | 12 | 16 | 12 | 16 |
| Hidden state size | 768 | 1024 | 768 | 1024 | 768 | 1024 | 768 | 1024 |

Table i. Selected properties of the different transformer models used (large models shaded).

|  | MODEL | | | | | | | | | | | |
|---|---|---|---|---|---|---|---|---|---|---|---|---|
|  | BERT$_{BASE}$ | | | GPT-2$_{SMALL}$ | | | XLNet$_{BASE}$ | | | RoBERTa$_{BASE}$ | | |
| **R-sq.** | *First* | *Last* | *Mean* | *First* | *Last* | *Mean* | *First* | *Last* | *Mean* | *First* | *Last* | *Mean* |
| Comb. | .668 | **.678** | .677 | .548 | **.630** | .611 | .655 | .660 | **.665** | .660 | .670 | **.673** |
| Best | .657 | **.671** | .667 | .520 | **.615** | .591 | .645 | .652 | **.657** | .647 | .652 | **.658** |

Table ii. Mean R-squared across all Binder features for different subword embedding approaches (first subword, last subword or mean across all subwords). Comb. = combined best layer per feature. Best = best single layer overall.

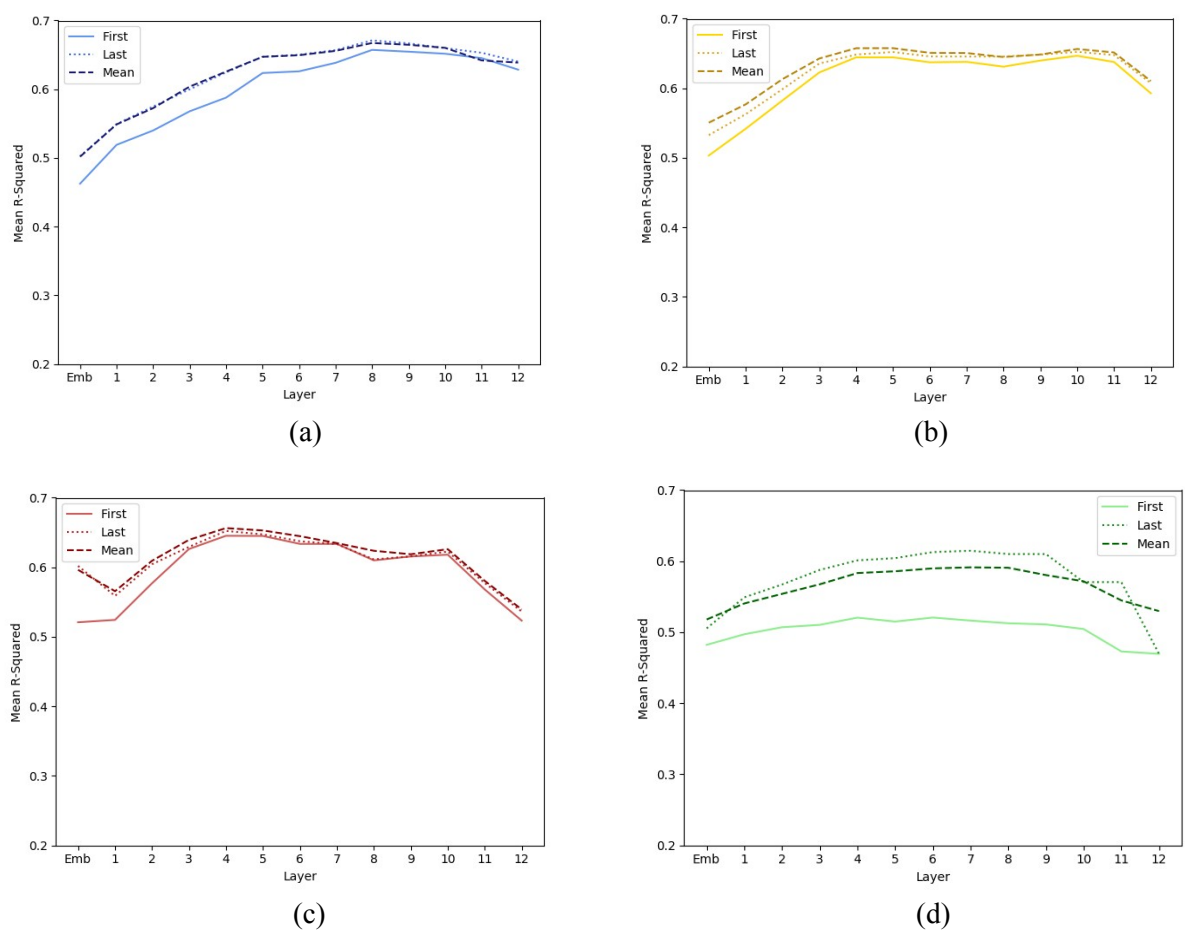

Figure i. performance of different subword embeddings across the 12 layers for (a) BERT$_{BASE}$ (b) RoBERTA$_{BASE}$ (c) XLNet$_{BASE}$ and (d) GPT-2$_{SMALL}$

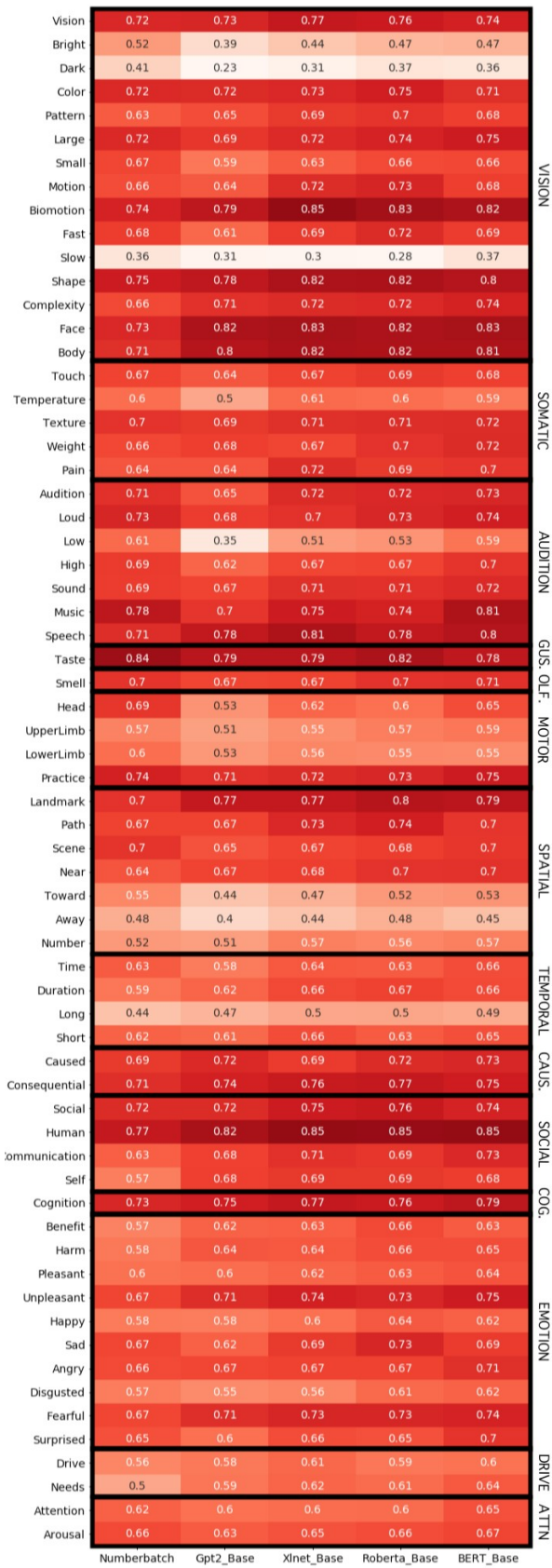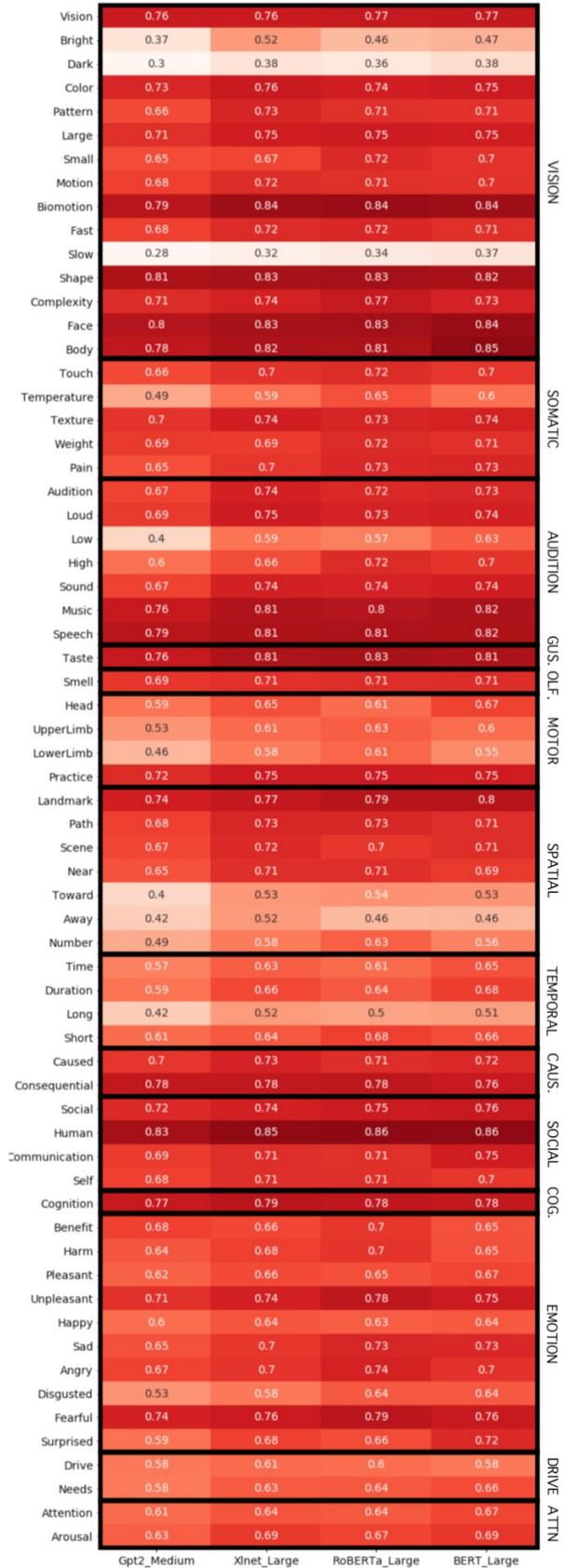

Figure ii. All feature R-squared scores for the Numberbatch baseline and (a) small models (b) large models, with Binder et al (2016) categories indicated.

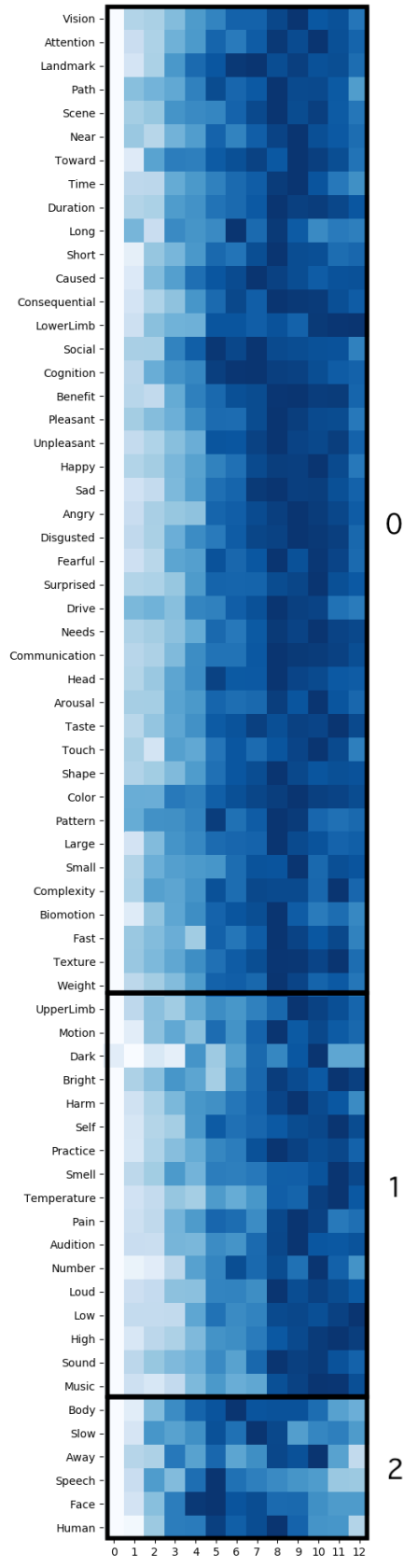

Figure iii. Rescaled R-squared scores across layers for different features in the BERT$_{BASE}$ model. K-means cluster memberships marked with black boxes.

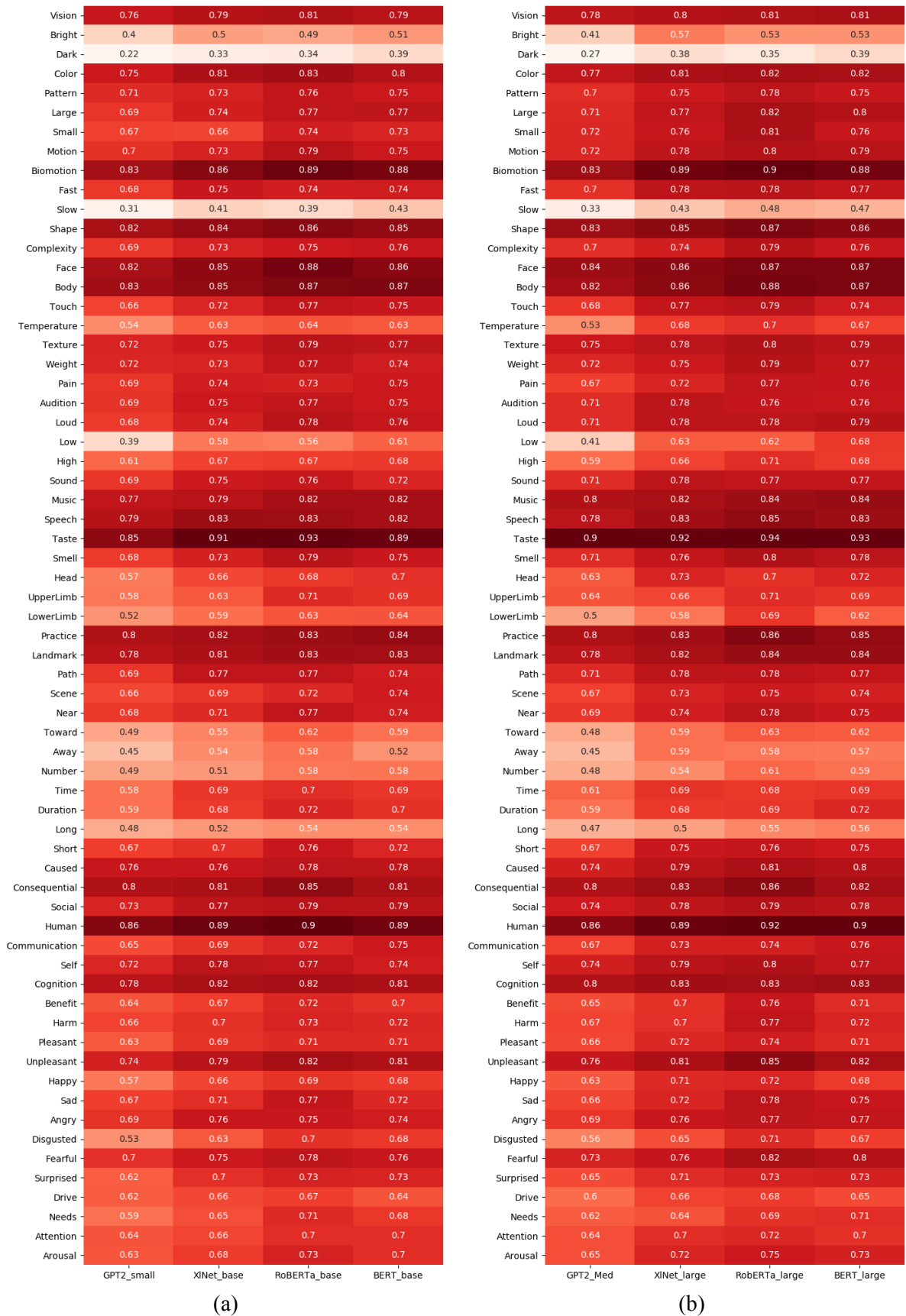

Figure iv. All feature R-squared scores for the (a) small and (b) large models for selected sentences of Experiment 1b.

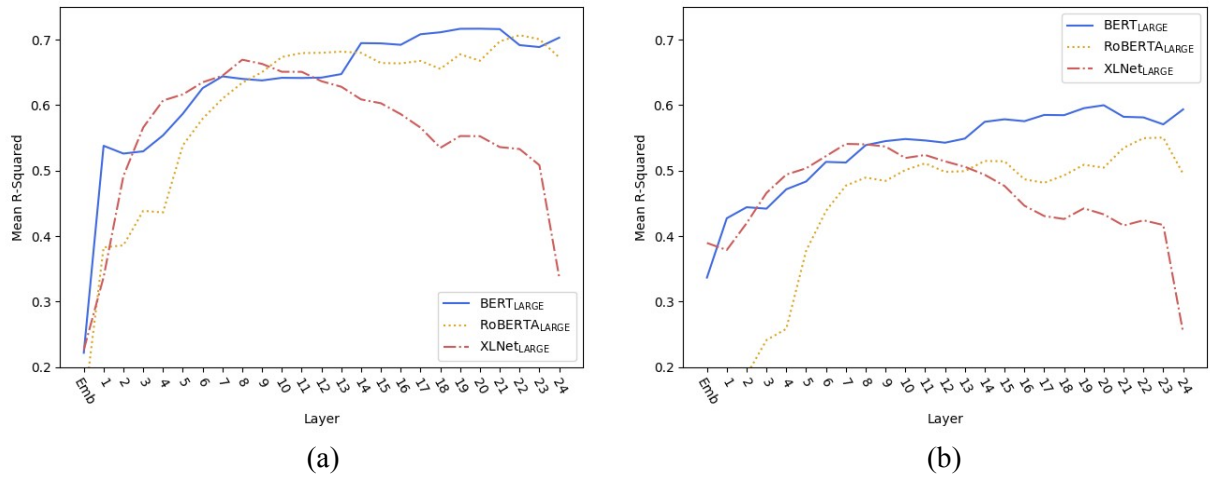

Figure v. Model per-layer mean R-squared scores for Experiment 2 using (a) individual word-pair property embedding and (b) mean across word-pairs property embedding.

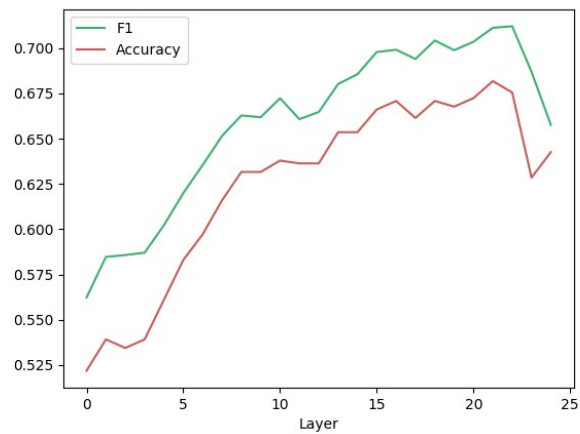

Figure vi. Raw BERT$_{LARGE}$ Accuracy and F1 scores on WiC dataset